\documentclass[10pt]{IEEEtran}
\usepackage{amsfonts}
\usepackage{amsmath}
\usepackage{amssymb}
\usepackage{bm}
\usepackage{cite}
\usepackage{graphicx}
\usepackage{epstopdf}
\usepackage{epsfig}
\usepackage{url}
\usepackage{setspace}
\usepackage{subfigure}
\usepackage{color}
\usepackage{algorithmicx}
\usepackage{algpseudocode}
\usepackage{amsmath}
\usepackage{algorithm,times}
\usepackage{threeparttable}
\usepackage{lineno}
\usepackage{multirow}
\usepackage{verbatim}
\usepackage{booktabs}
\graphicspath{{./figures/}}

\begin{document}

\begin{spacing}{1}
\title{Estimating the Direction and Radius of Pipe from GPR Image by Ellipse Inversion Model\thanks{%
The authors are with USTC-Birmingham Joint Research Institute in
Intelligent Computation and Its Applications, School of Computer Science
and Technology, University of Science and Technology of China, Hefei
230027, China (e-mail: zhou0612@ustc.edu.cn,  qqchern@ustc.edu.cn, saintfe@mail.ustc.edu.cn, hchen@ustc.edu.cn. Corresponding author: Huanhuan Chen).}
\author{Xiren Zhou, Qiuju Chen, Shengfei Lyu, Huanhuan~Chen,~\IEEEmembership{Senior Member,~IEEE}}
}
\maketitle

\begin{abstract}
Ground Penetrating Radar (GPR) is widely used as a non-destructive approach to estimate buried utilities. When the GPR's detecting direction is perpendicular to a pipeline, a hyperbolic characteristic would be formed on the GPR B-scan image. However, in real-world applications, the direction of pipelines on the existing pipeline map could be inaccurate, and it is hard to ensure the moving direction of GPR to be actually perpendicular to underground pipelines. In this paper, a novel model is proposed to estimate the direction and radius of pipeline and revise the existing pipeline map from GPR B-scan images. The model consists of two parts: GPR B-scan image processing and Ellipse Iterative Inversion Algorithm (EIIA). Firstly, the GPR B-scan image is processed with downward-opening point set extracted. The obtained point set is then iteratively inverted to the elliptical cross section of the buried pipeline, which is caused by the angle between the GPR's detecting direction and the pipeline's direction. By minimizing the sum of the algebraic distances from the extracted point set to the inverted ellipse, the most likely pipeline's direction and radius are determined. Experiments on real-world datasets are conducted, and the results demonstrate the effectiveness of the method.

\end{abstract}


\section{Introduction}
\label{introducation}

Underground pipeline are indispensable for the normal operation of urban cities, and part of pipelines are nearing their practical life and need to be replaced or repaired \cite{jaw2013locational}. To locate the buried utilities and revise the existing pipeline map, Ground Penetrating Radar (GPR) has been widely used due to its fast speed and minimal ground intrusion \cite{daniels2004ground}.

When detecting underground pipelines in an area, assuming that the underground medium is uniform (or little change), the permittivity of the underground medium could be estimated by fitting a hyperbolic feature generated by the pipeline on the GPR B-scan image\cite{shihab2005radius}, provided that the direction of the underground pipeline is perpendicular to the detecting direction, and the depth of the underground pipeline is within the effective detecting depth of the utilized GPR\cite{daniels2004ground}. Considerable efforts have been devoted to extract and fit hyperbolas on B-scan images \cite{capineri1998advanced, porrill1990fitting,caorsi2005electromagnetic,maas2013using,chen2010robust,dou2016real}, 
In \cite{chen2010probabilistic}, a probabilistic hyperbola mixture model is proposed to estimate buried pipelines from GPR B-scan images, where the Expectation-Maximization (EM) algorithm is upgraded to extract multiple hyperbolas, and a designed fitting algorithm is then adopted to fit these hyperbolas.
In\cite{dou2016real}, Dou \emph{et al.} proposed the Column-Connection Clustering (C3) algorithm. The algorithm could extract point clusters from B-scan image, which are then identified by a neural-network-based method to locate hyperbolic ones.
In our previous work\cite{zhou2018automatic}, a GPR B-scan image interpreting model has been proposed to extract and fit hyperbolic signatures on B-scan images, estimate the permittivity of the detected area, and obtain the information of buried pipelines. Experiments in different media has validated the effectiveness of the model. By applying these methods, the permittivity of the detected area could be roughly obtained. 

After obtaining the permittivity of the detected area through above methods, the ordinate of the GPR B-scan image can be converted from time to depth\cite{shihab2005radius}. When detecting other pipelines in this area, their depth could be inferred directly from the GPR B-scan image\cite{zhou2018automatic}. To revise the existing pipeline map, the follow-up work is to further estimate the direction and the radius of each pipeline from GPR data. When the pipeline's direction is not perpendicular to the GPR's detecting direction, the cross section of the pipeline in the GPR's detecting direction is elliptical, and the generated feature is not hyperbolic\cite{daniels2004ground}. In this case, fitting the generated feature by hyperbolic equation would cause errors. In \cite{chen2011buried}, the direction of the underground pipeline is roughly estimated by the statutory records of buried ultities, and  detections at different directions are then conducted to derive the specific direction of each pipeline. In \cite{dou20163d}, a Marching-Cross-Sections (MCS) algorithm is proposed to merge the individual hypothesized pipeline segments. In this algorithm, parallel scan-lines are established, and the Kalman Filter (KF) \cite{jazwinski2007stochastic} is extended to connect hypothesized points on the pipe, and infer the direction and position of each pipeline. The above-mentioned methods determine the direction of each pipeline by multiple detections in different directions, or connecting two or more detected positions of a pipeline. In addition, when the pipeline's direction is not perpendicular to the GPR's detecting direction, it could be inaccurate to measure the radius of the pipeline by fitting hyperbolic signatures on the B-scan image. 

In this paper, the Ellipse Inversion Model is proposed to estimate the direction and radius of pipeline from GPR B-scan image and revise the existing pipeline map. The model consists of two parts: GPR B-scan image processing and Ellipse Iterative Inversion Algorithm (EIIA). The GPR B-scan image is firstly processed with downward-opening point set extracted by extending part of our previous work\cite{zhou2018automatic}. As the angle between the GPR's detecting directions and the pipe direction might not be perpendicular, the cross section of the pipeline could be elliptical. In this case, the EIIA iteratively inverts the extracted downward-opening point set to the elliptical cross section of the pipe. By minimizing the sum of the algebraic distances from these points to the inverted ellipse, the most likely pipe direction and radius are determined.

This rest of this paper is organized as follows. The GPR B-scan
image processing is discussed in Section II. Section III provides the Ellipse Iterative Inversion Algorithm. Experiments are conducted and analyzed in Section IV. Finally, conclusions are drawn in Section V.

\section{GPR B-scan image processing}
In this section, the downward-opening signature on GPR B-scan images generated from the elliptical cross section is analyzed. Then the method to extract point set with downward-opening signatures is introduced.

\subsection{The downward-opening signatures generated by the elliptical cross section}

\begin{figure}[htbp]
	\centering
	\includegraphics[width=0.32\textwidth]{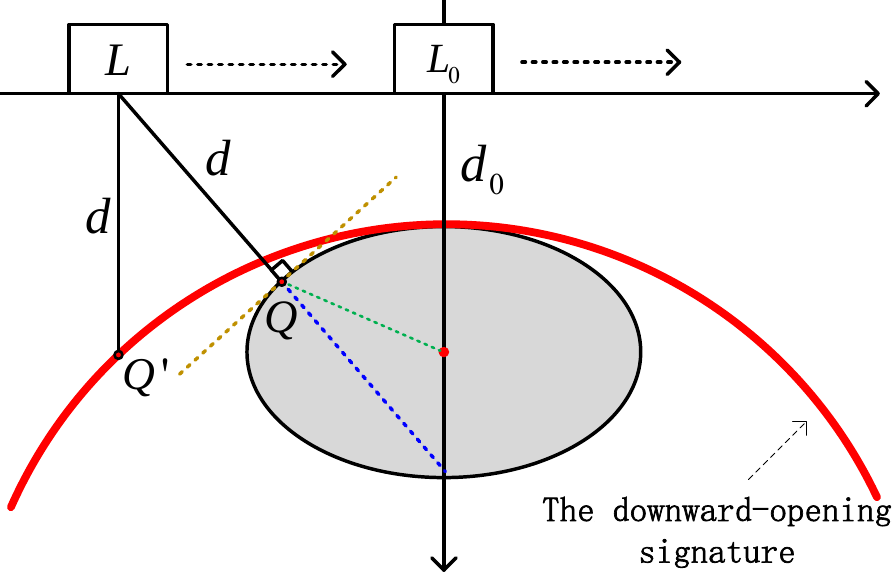}
	\caption{The case where the cross section of the pipeline in the detecting direction of the GPR is elliptical. The gray ellipse indicates the cross section of the pipe. $L$ is the location of the GPR, and $L_0$ means the location where the GPR is directly above the pipeline. $Q$ is the closest point on the ellipse to the GPR position $L$, and the distance from $L$ to the ellipse is $|LQ|=d$. The red line indicates the downward-opening signature generated by the pipeline on the GPR B-scan image. $Q'$ is the point on the downward-opening signature directly below $L$ generated by $Q$, and $|LQ|=|LQ'|=d$.}
	\label{ell1}
\end{figure}

Fig. \ref{ell1} illustrates the schematic diagram when the GPR's detecting direction is not perpendicular to the pipeline's direction, where the cross section of the pipeline is elliptical. 
It could be seen that when the GPR is at $L$, the extension line (blue dotted line) of $LQ$, which indicates the distance $d$ between the GPR and the ellipse, does not pass through the center\footnote{The center of the ellipse is the midpoint between the two focal points of the ellipse.} of the ellipse, as the blue and the green dashed lines do not coincide in Fig. \ref{ell1}. Therefore, the feature produced by the elliptical cross section on the GPR B-Scan image could not be described by the hyperbolic equation \cite{shihab2005radius}.
As GPR moves from $L$ to $L_0$, the distance between it and the pipeline gradually decreases, and the trajectory when the GPR moves gradually away from $L_0$ is symmetrical, thus the feature produced by the elliptical cross section on the GPR B-Scan image could be represented by the red downward-opening curve in Fig. \ref{ell1}.

\subsection{Extract point set with downward-opening signatures}

In our previous work\cite{zhou2018automatic}, the GPR B-scan image preprocessing method and Open-Scan Clustering Algorithm (OSCA) have been proposed. In this paper, the preprocessing method firstly eliminates the discrete noises and transforms the original image into binary image. The obtained binary image is then scanned by OSCA from top to bottom with downward-opening point clusters identified and extracted. 

\begin{figure}[htbp]
	\centering
	\subfigure[]{ \centering
		\label{Bscan31}
		\includegraphics[width=0.2\textwidth]{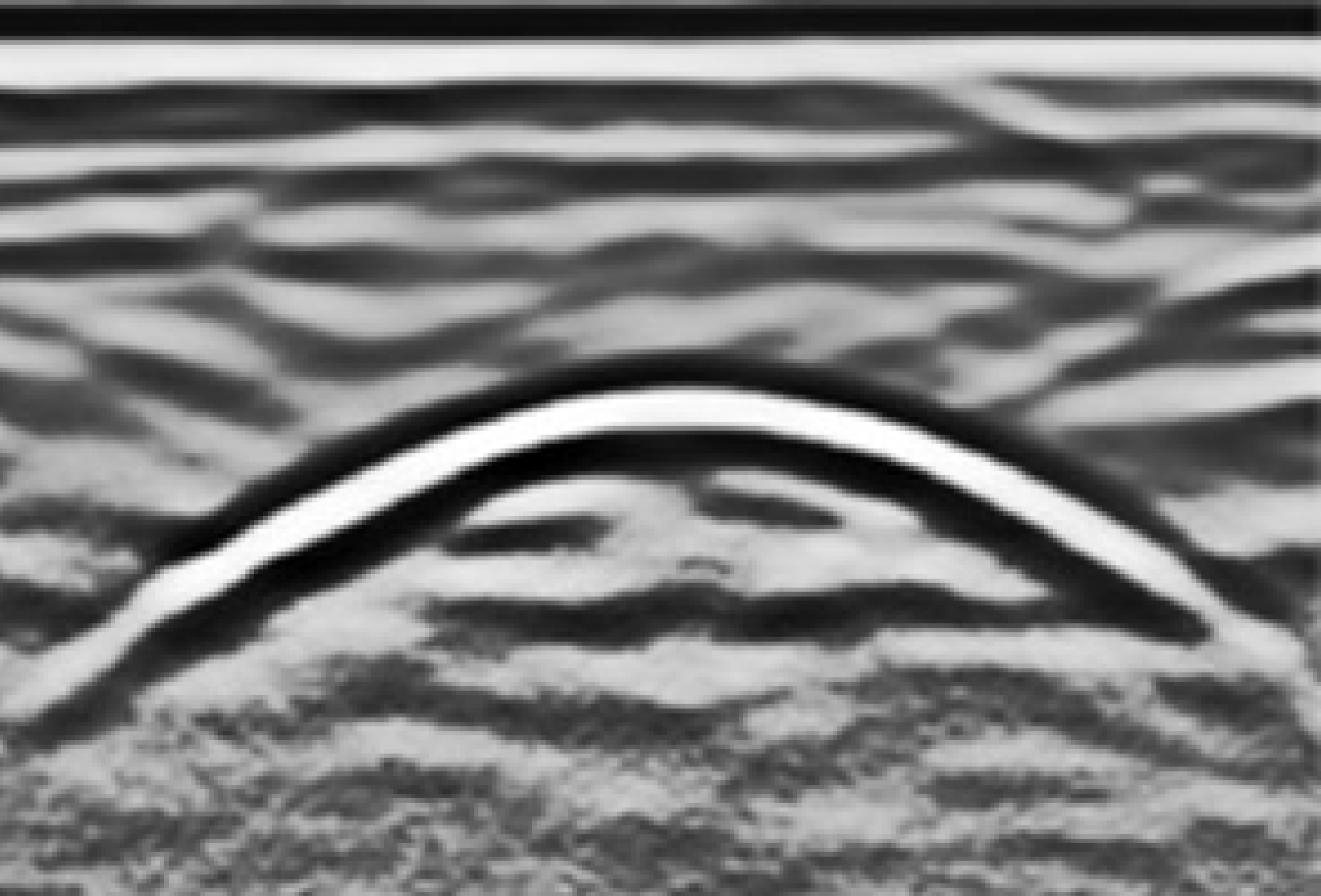}}
	\subfigure[]{ \centering
		\label{Bscan32}
		\includegraphics[width=0.2\textwidth]{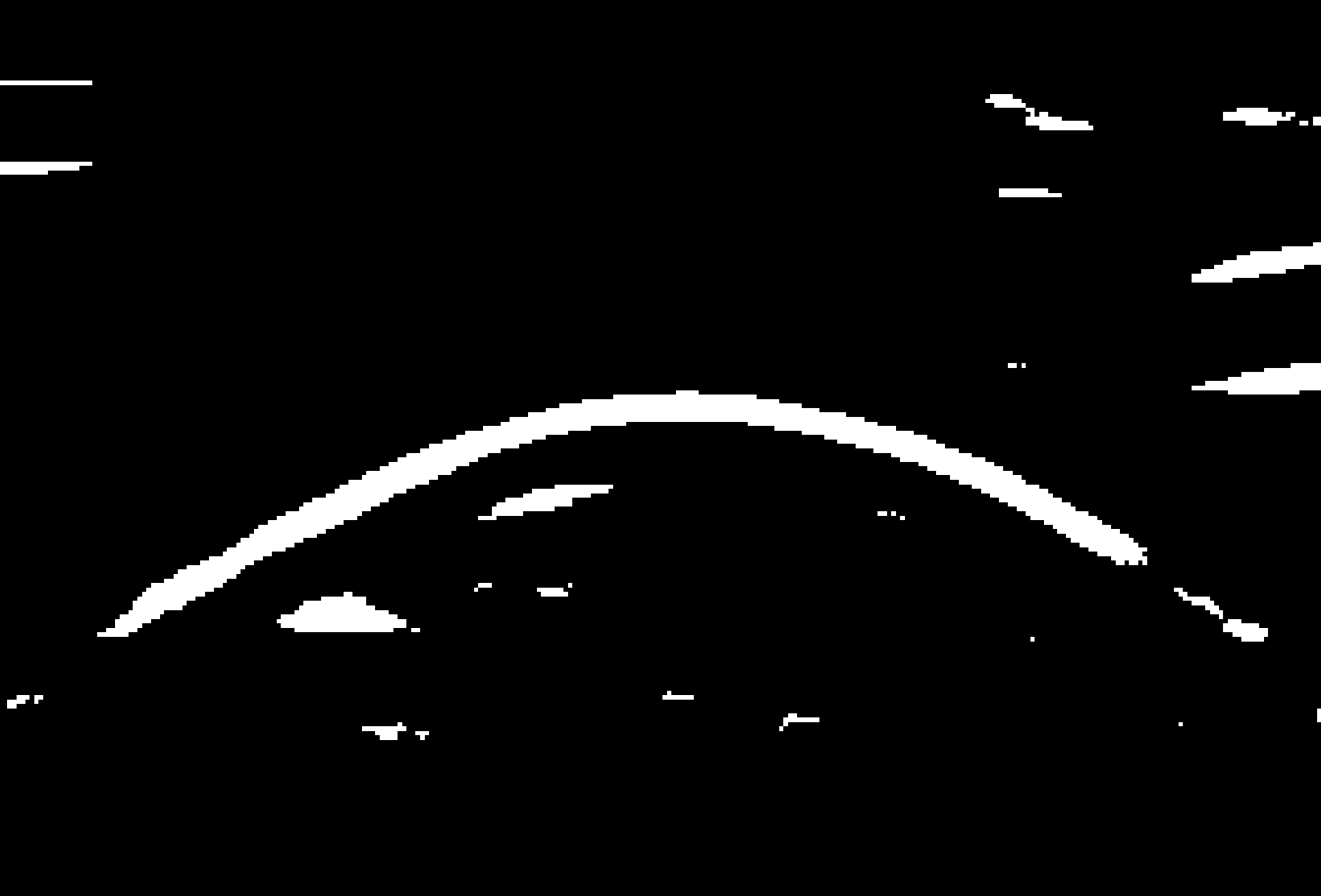}}
	\subfigure[]{ \centering
		\label{Bscan33}
		\includegraphics[width=0.2\textwidth]{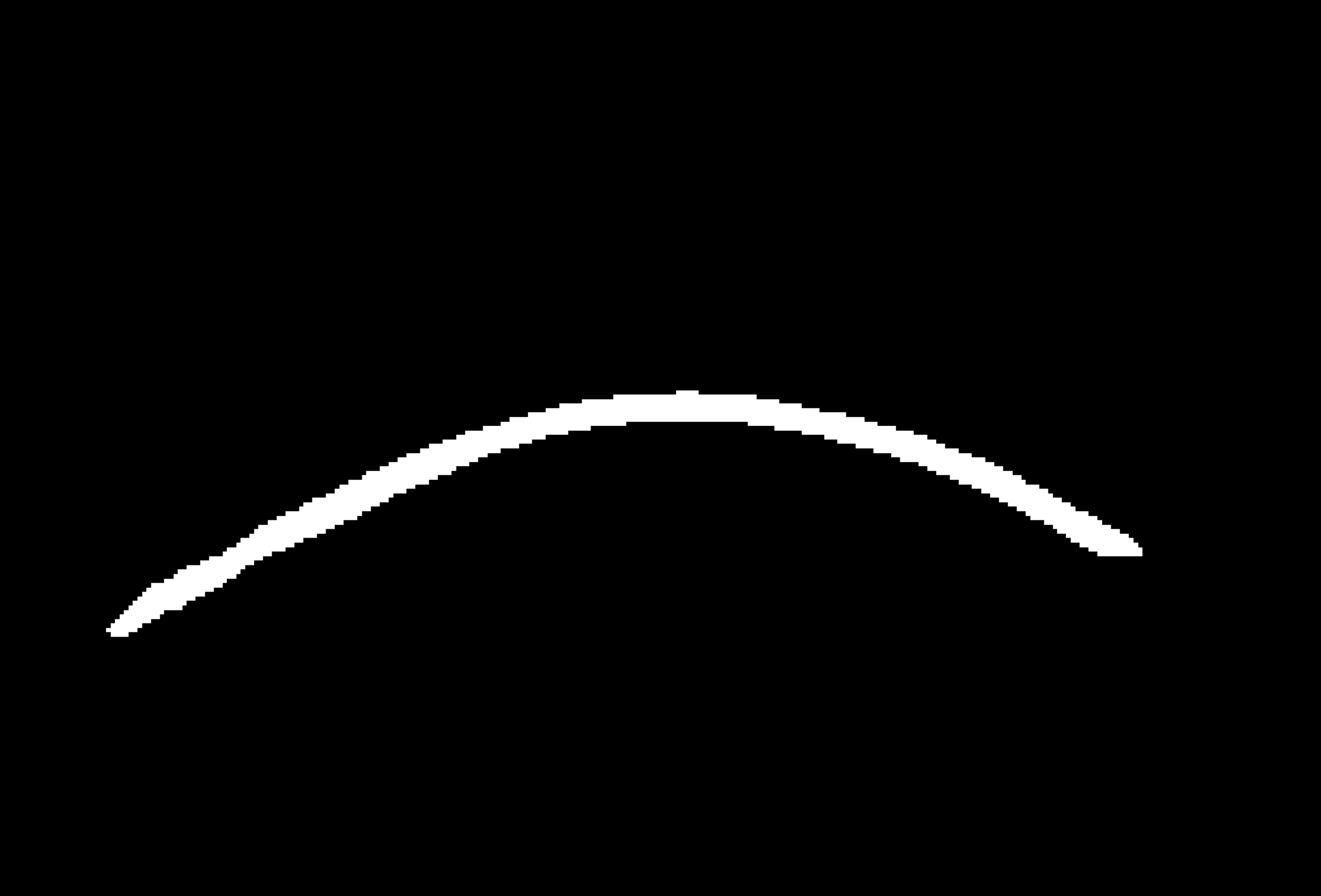}}
	\subfigure[]{ \centering
		\label{Bscan34}
		\includegraphics[width=0.2\textwidth]{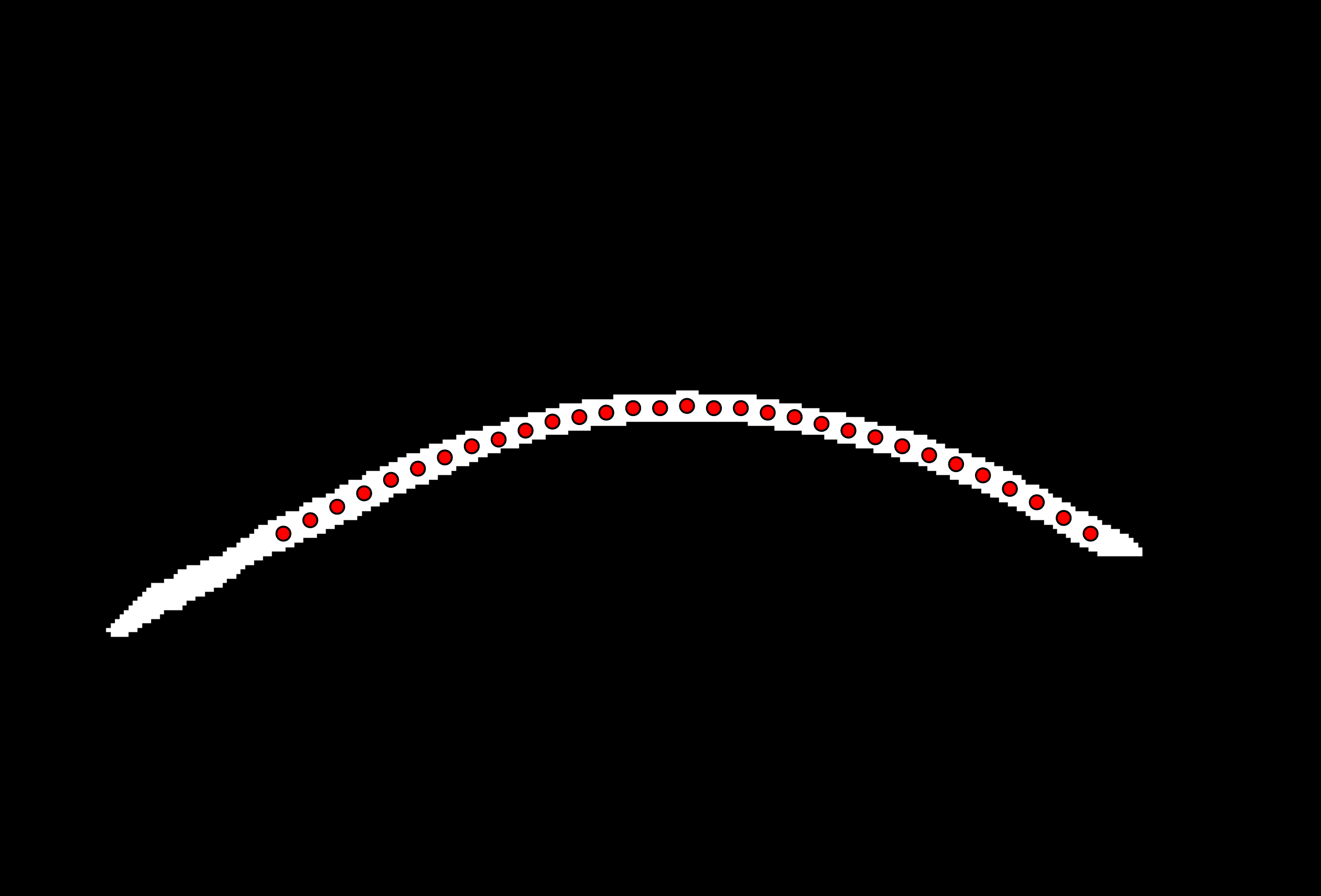}}
	\caption{The processing flow of extracting a downward-opening point set from a GPR B-scan image. (a) is the original image. (b) is the preprocessed binary image. (c) is the obtained result after OSCA. (d) shows the extracted downward-opening point set.}\label{bscanpro}
\end{figure}

Since the top of the pipe is closest to the surface and the generated feature on the image is also the most obvious, the column of the identified point cluster that is closest to the ground are selected with several columns on the left and right sides. The distance between each two adjacent selected columns is set to be $2$cm in this paper, and the midpoints of these columns are extracted to form a point set $\mathbf{P}$ as
\begin{equation}
	\label{pointset}
	\mathbf{P}=\left \{ P_i\left ( x_{i},y_{i} \right ) |\ 0\leq i\leq n\right \}
\end{equation}
which is further inverted to the elliptical cross section of the buried pipe in the next section.  
Fig. \ref{bscanpro} illustrates the process of extracting a downward-opening point set from an original GPR B-scan image.

\section{Ellipse Iterative Inversion Algorithm}
The EIIA aims to revert the downward-opening point set $\textbf{P}$ to the elliptical cross section of the pipe, which consists of two parts: ellipse fitting algorithm, and updating $\mathbf{P}$ by converting the coordinates of each point. The procedure of EIIA is presented at the end of this section.

\subsection{Ellipse fitting algorithm}

The fitting of a general conic can be approached by minimizing the sum of squared algebraic distances $\textit{D}\left (\mathbf{P} \right )$ of the curve\cite{gander1994least} to the point set $\mathbf{P}=\left \{ \left ( x_{i},y_{i} \right ) |1\leq i\leq n\right \}$  as 

\begin{equation}
	\text{Minimize}\  \textit{D}\left ( \mathbf{P} \right )=\sum_{i=1}^{n}\textit{F}\left (\mathbf{A},\mathbf{x_{i}} \right )^2,
\end{equation}
\begin{equation}
	\label{conic}
	\textit{F}\left ( \mathbf{A}, \mathbf{x} \right )=\mathbf{A}\cdot \mathbf{x}=Ax^{2}+Bxy+Cy^{2}+Dx+Ey+F.
\end{equation}
where $ \mathbf{A}=\left [ A,B,C,D,E,F \right ]^{T}$, $\mathbf{x}=\left [ x^{2},xy,y^{2},x,y,1 \right ]$, and $\textit{F}\left ( \mathbf{A},\mathbf{x} \right )$ is the ``algebraic distance'' of a point $\left ( x,y \right ) $ to the conic $\textit{F}\left (\mathbf{A},\mathbf{x} \right )=0$. 

In our model, an ellipse could be presented as
\begin{equation}
	\label{ellp1}
	\frac{\left ( x-x_{0} \right )^2}{a^2}+\frac{\left ( y-y_{0} \right )^2}{b^2}=1.
\end{equation}
By transforming Equation \eqref{ellp1} into the form of Equation \eqref{conic}, the following equation is obtained
\begin{equation}
	\label{ellp2}
	\begin{aligned}
		b^{2} x^{2}&+a^{2}y^{2}-2b^{2}x_{0}x-2a^{2}y_{0}y\\
		&+\left ( a^{2}y_{0}^{2}+b^{2}x_{0}^{2}-a^{2}b^{2} \right )=0.
	\end{aligned}
\end{equation}
Comparing Equations \eqref{ellp2} and \eqref{conic}, it could be seen that $B=0 $. To ensure the fitted conic to be elliptical, $4AC=1$ is utilized as the constraint to limit the fitted curve to be an ellipse. Therefore, fitting the point set $\mathbf{P}$ to an ellipse could be formulated as 
\begin{equation}
	\begin{aligned}
		\text{Minimize}\ \textit{D}&\left ( \mathbf{P} \right )=\sum_{i=1}^{n}\textit{F}\left (\mathbf{A},\mathbf{x_{i}} \right )^2\\
		\ \text{s.t.}\ & B=0,\  4AC=-1,
	\end{aligned}
\end{equation}
which is a convex optimization (CVX) problem and could be solved by the method proposed in\cite{grant2008graph}.

\subsection{Updating the coordinates of each point in $\mathbf{P}$}
Given the point set $\mathbf{P}=\left \{ P_i\left ( x_{i},y_{i} \right ) |\ 0\leq i\leq n\right \}$ as Equation \eqref{pointset}, the projection of each point $P_i=(x_i,y_i)$ on the $X$ axis is $P_{i,X}$, which is the position of the GPR that collects the signal at $P_i$. 
\begin{figure}[htbp]
	\centering
	\subfigure[]{ \centering
		\label{rotate1}
		\includegraphics[height=1.35in]{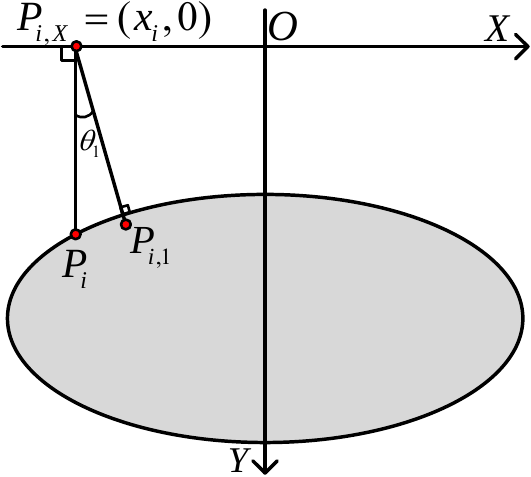}}
	\subfigure[]{ \centering
		\label{rotate2}
		\includegraphics[height=1.35in]{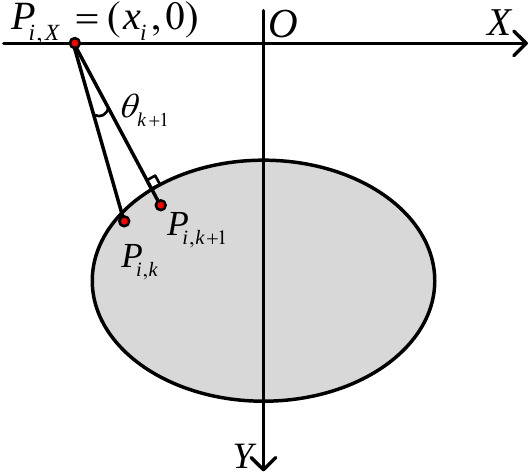}}
	\caption{(a) and (b) show the first and the $(k+1)$th coordinate conversion of $P_i$. The gray ellipse represents the elliptical pipe section fitted before coordinate conversion.}\label{rotatepoint}
\end{figure}
As Fig. \ref{rotate1} shows, $\mathbf{P}$ is firstly fitted into an ellipse. Then $P_i$ rotates around the point $P_{i,X}$ to $P_{i,1}$ with the angle of $\theta_1$, where $|P_{i,X}P_{i}|=|P_{i,X}P_{i,1}|$ and the distance between $P_{i,X}$ and the fitted ellipse is the shortest, that is, the straight line determined by $P_{i,X}$ and $P_{i,1}$ is perpendicular to the tangent line at the closest intersection of the line and the fitted ellipse. The coordinates of $P_{i,1}$ could be obtained by  
\begin{equation}
	\left\{
	\begin{aligned}
		&x_{i,1}=x_i-y_i\sin\theta_1,\\
		&y_{i,1}=y_i\cos\theta_1.
	\end{aligned}
	\right.
\end{equation}
where $\theta_1$ could be obtained by calculating the shortest distance from a $P_{i,X}$ outside the ellipse to the ellipse\cite{uteshev2018point}. The rotation is applied on every point in $\mathbf{P}$, and the coordinate of these points are updated, by which $\mathbf{P}$ is updated to $\mathbf{P}_1=\left \{ P_{i,1}\left ( x_{i,1},y_{i,1} \right ) |\ 0\leq i\leq n\right \}$.

The ellipse fitting and rotation are alternately and iteratively performed on $\mathbf{P}$. As Fig. \ref{rotate2} shows, after $k$th iteration, the coordinates of $P_{i,k+1}$ at the $k+1$ iteration are updated as
\begin{equation}
	\label{update2}
	\left\{
	\begin{aligned}
		&x_{i,k+1}=x_{i}-y_{i}\sin (\sum_{j=1}^{k+1}\theta_j),\\
		&y_{i,k+1}=y_{i}\cos (\sum_{j=1}^{k+1}\theta_j).
	\end{aligned}
	\right.
\end{equation}
where $\theta_{j}$ is the $j$th rotating angle. The condition for stopping the above iteration of fitting and rotation is that the iterative number reaches the set threshold $K$, or the sum of algebraic distance $\textit{D}\left ( \mathbf{P} \right )$ from each point in $\mathbf{P}$ to the fitted ellipse and tends to be stable.

\subsection{The Procedure of Ellipse Iterative Inversion Algorithm}
Based on the above ellipse fitting algorithm and coordinate updating method, the procedure of the Ellipse Iterative Inversion Algorithm is presented in the following:
\begin{enumerate}
	\item (Input) Point set $\mathbf{P}=\left \{ P_i\left ( x_{i},y_{i} \right ) |\ 0\leq i\leq n\right \}$ extracted from the GPR B-scan image; the maximum number of iterations $K$; the threshold of the sum of algebraic distance $D_t$.
	\item (Ellipse Fitting) Fit $\mathbf{P}$ into the ellipse by the proposed ellipse fitting algorithm, along which the sum of algebraic distance $\textit{D}\left ( \mathbf{P} \right )$ from each point in $\mathbf{P}$ to the fitted ellipse is obtained
	\item (Coordinate Updating) Updating the coordinates of each point in $\mathbf{P}$ by Equation \eqref{update2}.
	\item (Check for convergence) If the iterative number reaches $K$, or $\textit{D}\left ( \mathbf{P} \right )\leq D_t$, output the ellipse equation with the smallest $\textit{D}\left ( \mathbf{P} \right )$. Otherwise, return to Step 2 (Ellipse Fitting).   
\end{enumerate}

\begin{figure}[htbp]
	\centering
	\subfigure[]{ \centering
		\label{angle}
		\includegraphics[height=1.1in]{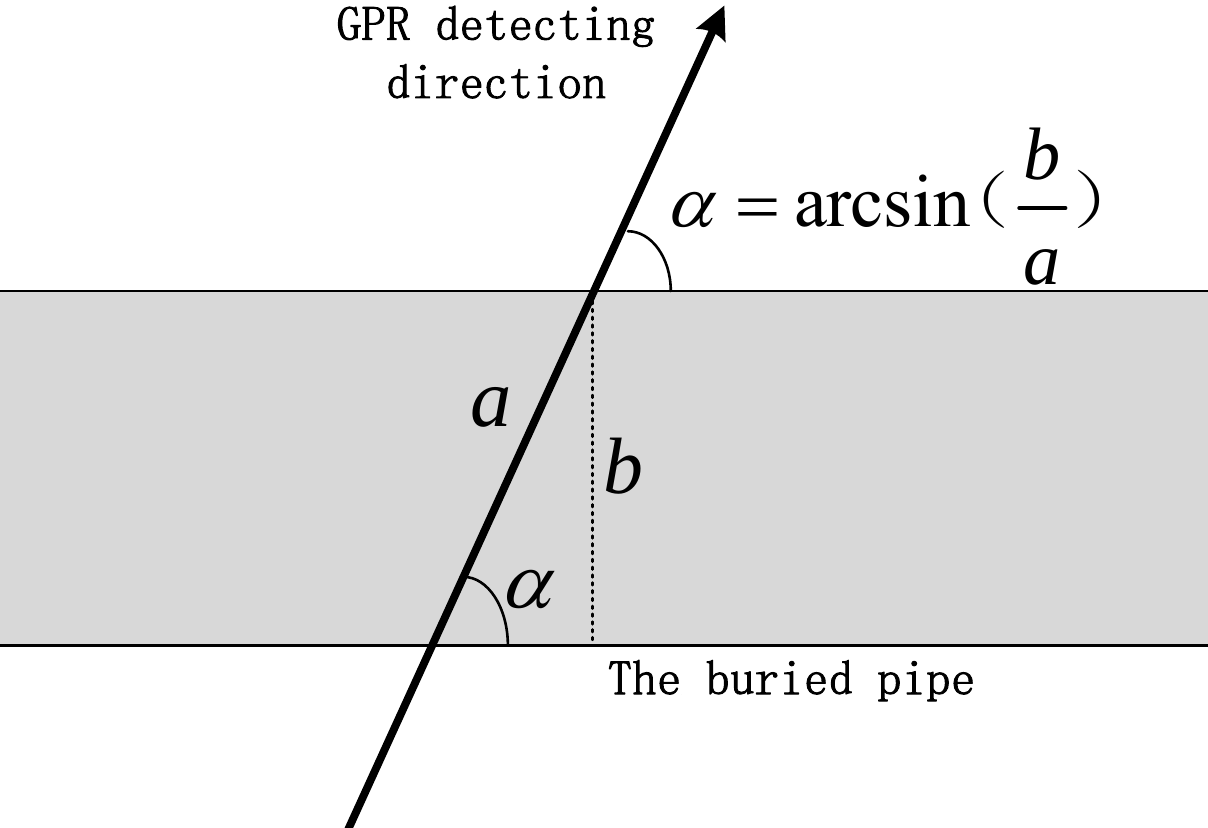}}
	\subfigure[]{ \centering
		\label{pipedirection}
		\includegraphics[height=1.1in]{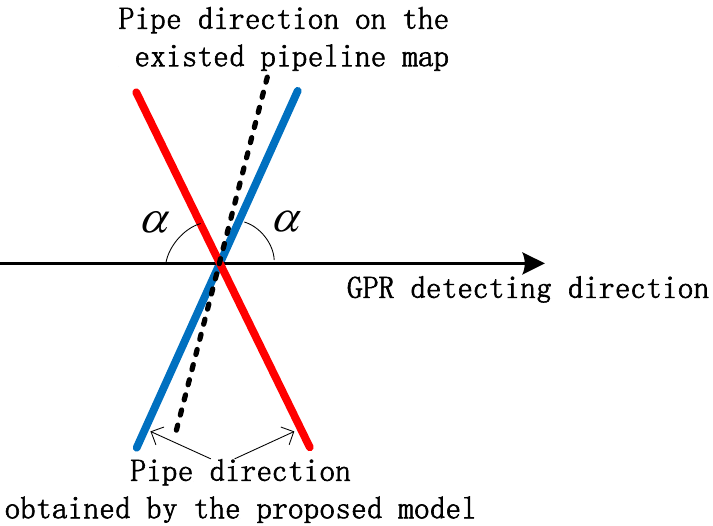}}
	\caption{(a) The angle $\alpha$ between the pipe and the GPR's detecting direction could be calculated as $\textit{arcsin}\frac{b}{a}$, where $a$ and $b$ are the parameters of obtained ellipse equation as Equation \eqref{ell1}.  (b) The black line with an arrow indicates the direction of GPR, the blue and red lines represent two possible pipeline's directions, both of which have an angle of $\alpha$ with the GPR's detecting direction. The dashed line indicates the direction of the pipe on the pipeline map. The blue line is adopted as the pipeline's direction, since it has smaller angle with the black dashed line compared with the red line.}	\label{3}
\end{figure}

The angle $\alpha$ between the pipeline and the GPR's detecting direction could be calculated as $\textit{arcsin}\frac{b}{a}$, and $b$ indicates the radius of the pipe. $a$ and $b$ are the parameters of obtained ellipse equation as Equation \eqref{ell1}, which are shown in Fig. \ref{angle}. 
As Fig. \ref{pipedirection} shows, for an $\alpha$, there are two possible pipeline's directions as the red and blue line. Therefore, when choosing the detecting direction of GPR, we would choose a direction that is not perpendicular to the pipeline according to the existing pipeline map, such as $\frac{4}{9}\pi$ or $\frac{3}{8}\pi$. The obtained pipeline's direction that has a smaller angle with the direction on the existing pipeline map is adopted as the modified pipeline's direction, as the blue line in Fig. \ref{pipedirection}.

\section{Experimental study}
To evaluate the effectiveness of the proposed model, real-world experiments are conducted in this section. After that, the analysis of the experimental results are presented.

\subsection{Experiments on real-world datasets}

\begin{figure}[htbp]
	\centering
	\subfigure[]{ \centering
		\label{host}
		\includegraphics[height=0.85in]{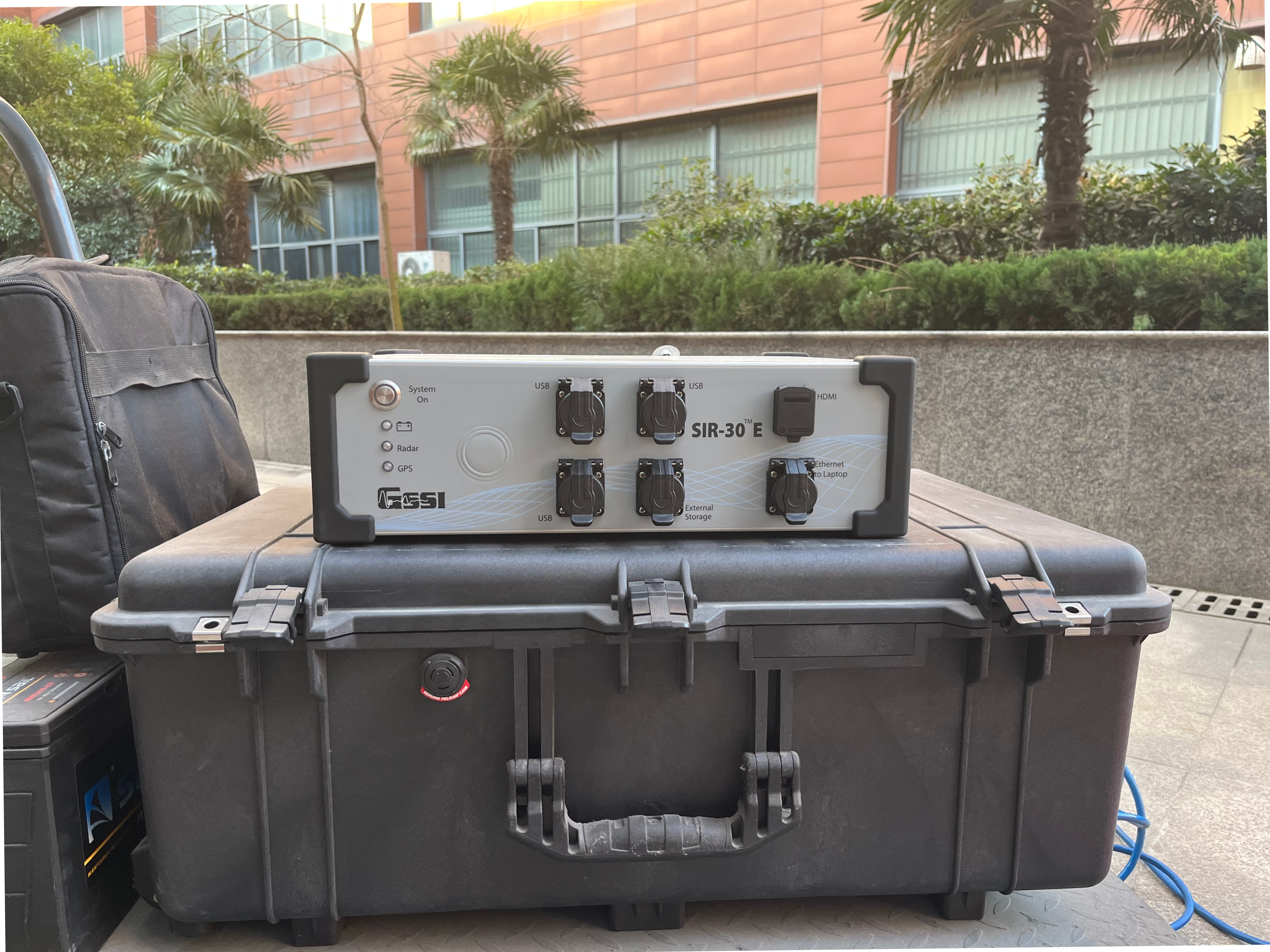}}
	\subfigure[]{ \centering
		\label{200mhz}
		\includegraphics[height=0.85in]{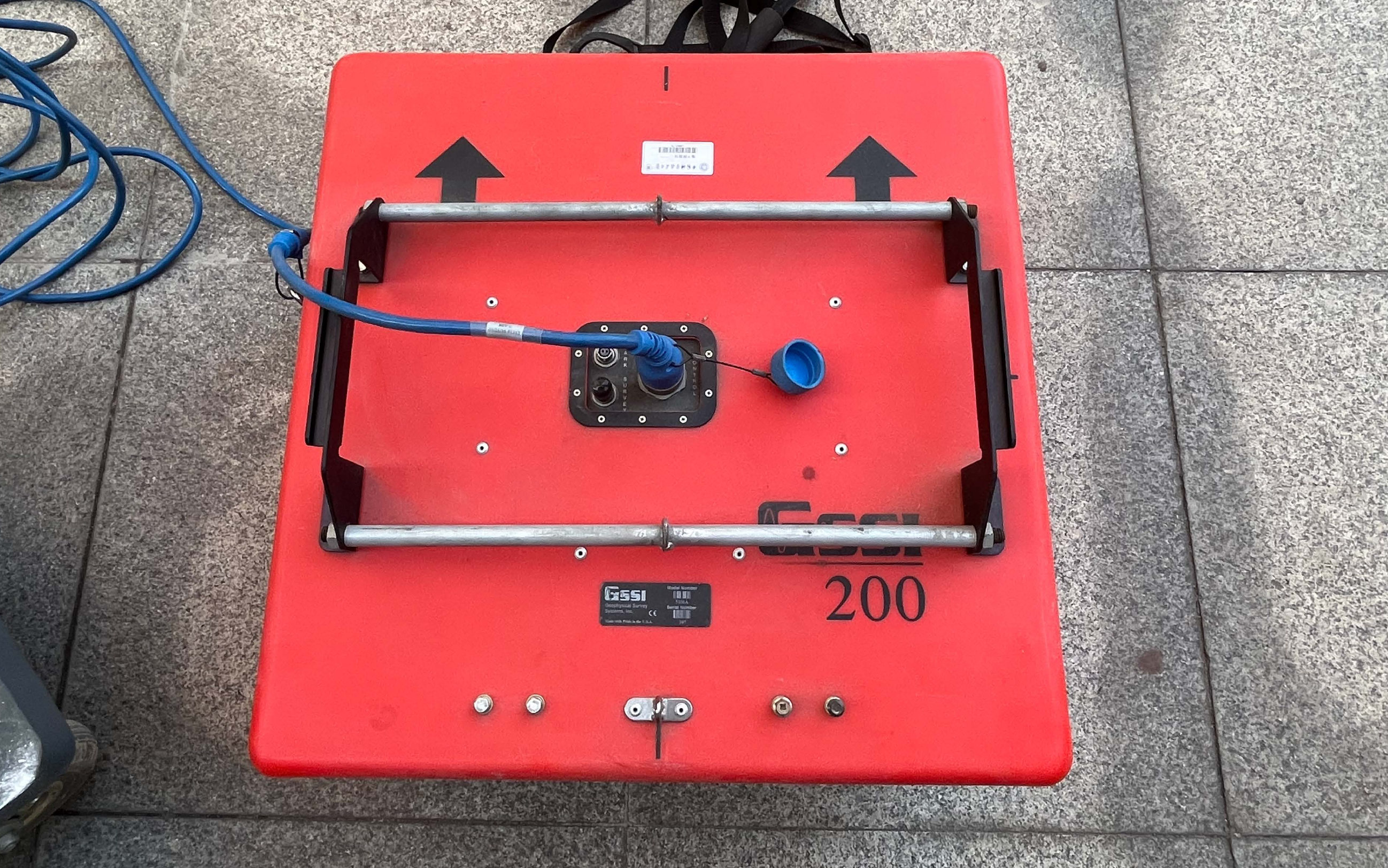}}
	\caption{(a) and (b) are the host and 200MHz antenna of the utilized GSSI SIR-30 GPR.}	\label{ugpr}
\end{figure}

\begin{figure}[htbp]
	\centering
	\subfigure[]{ \centering
		\label{area1}
		\includegraphics[height=1.05in]{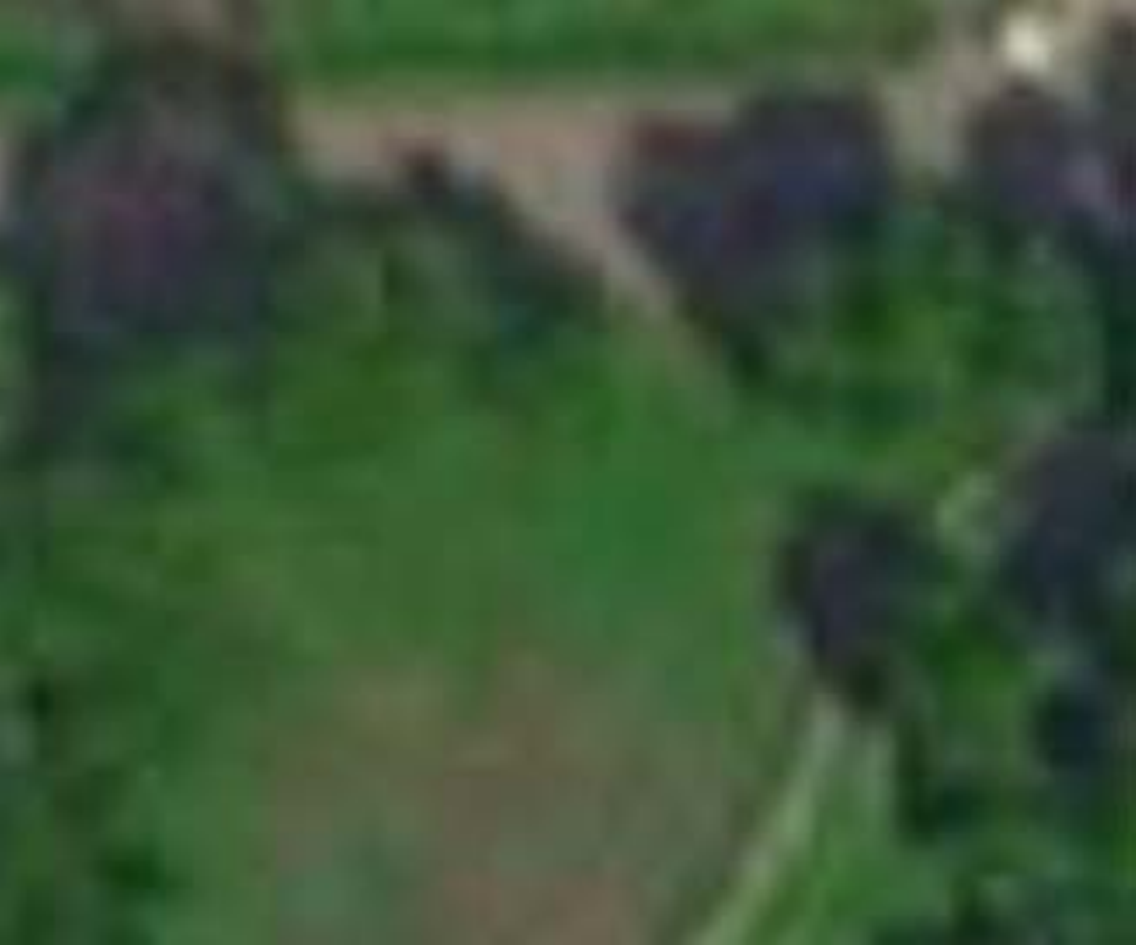}}
	\subfigure[]{ \centering
		\label{area2}
		\includegraphics[height=1.05in]{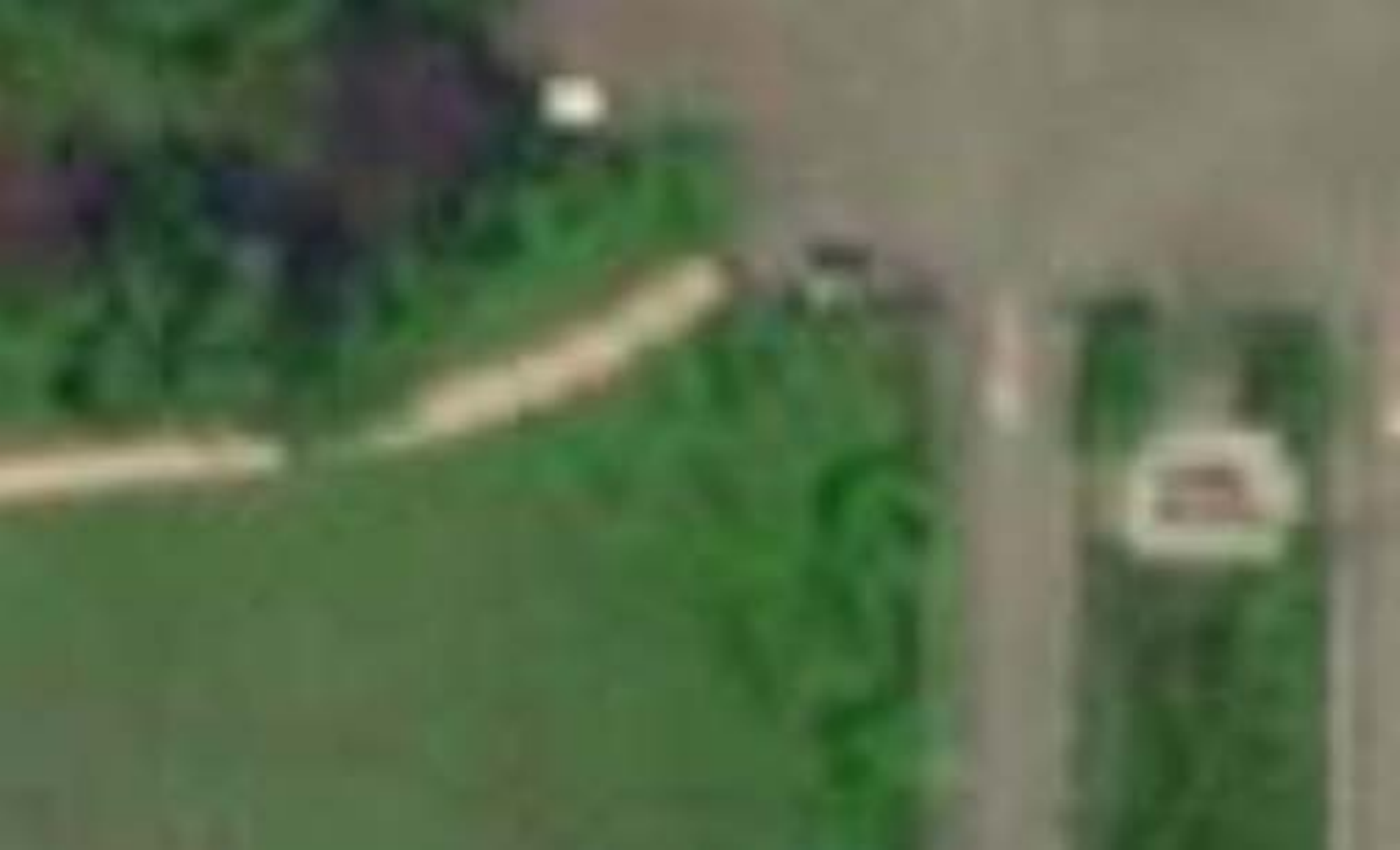}}
	\subfigure[]{ \centering
		\label{area3}
		\includegraphics[height=1.1in]{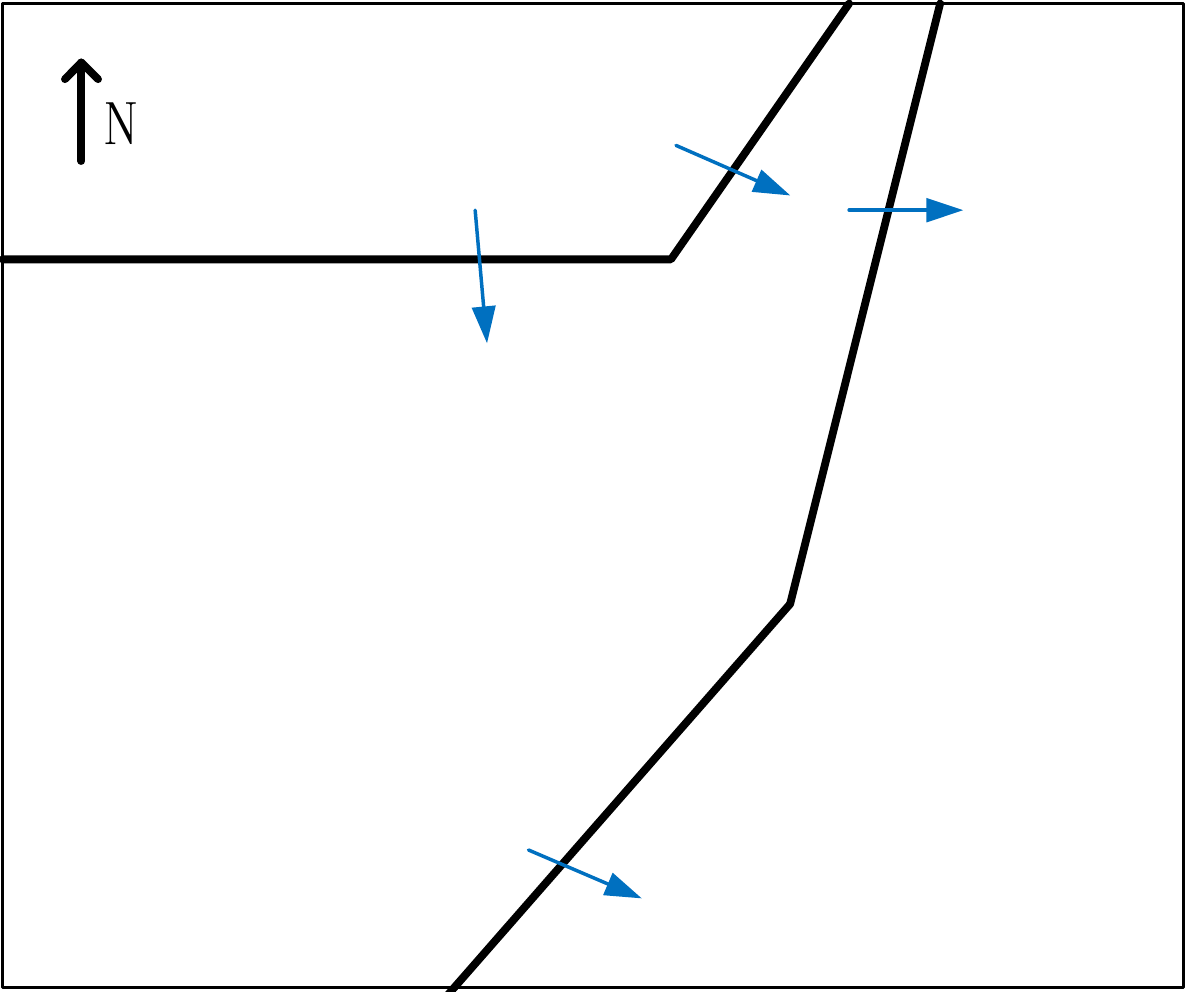}}
	\subfigure[]{ \centering
		\label{gssi2}
		\includegraphics[height=1.1in]{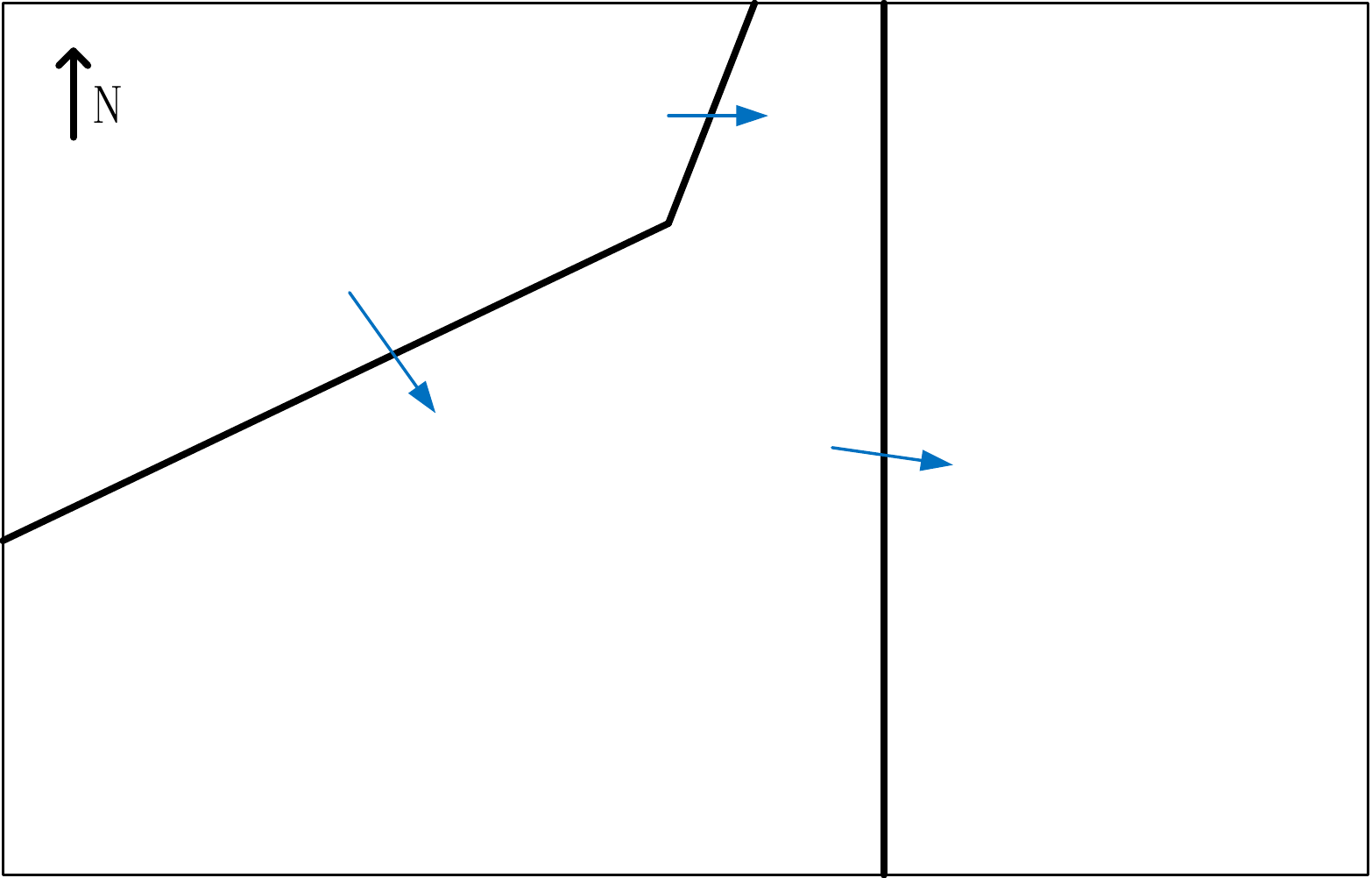}}
	\caption{The two selected areas are shown as (a) and (b). The existing pipeline maps are shown in (c) and (d), where the black lines indicate the pipeline. The blue lines with arrows indicate the GPR's detecting direction and position.}	\label{area_and_map}
\end{figure}

Two experimental areas are identified, where pipeline excavation and repair work have been carried out within two months, but the latest pipeline maps are revised one year ago. 
GSSI’s SIR-30 GPR with 200MHz antenna is utilized to collect GPR B-scan images. The utilized GPR is presented in Fig. \ref{ugpr}. The two selected areas with existing pipeline maps are shown in Fig. \ref{area_and_map} and the selected detecting positions and directions are also presented.

\begin{figure}[htbp]
	\centering
	\subfigure[]{ \centering
		\label{gprdirection21}
		\includegraphics[width=0.20\textwidth]{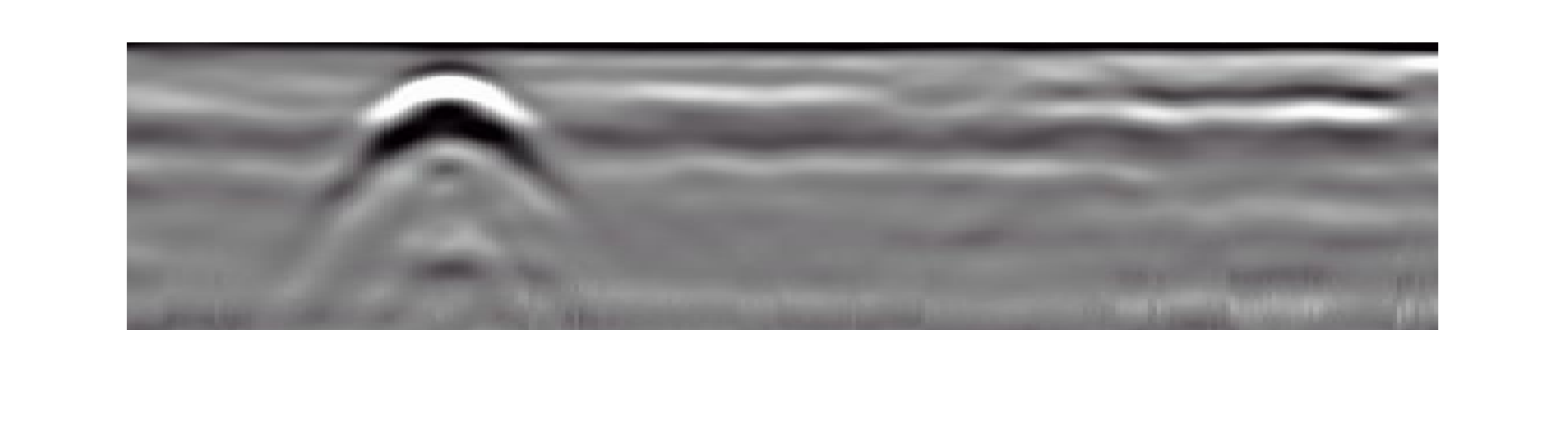}}
	\subfigure[]{ \centering
		\label{gprdirection22}
		\includegraphics[width=0.20\textwidth]{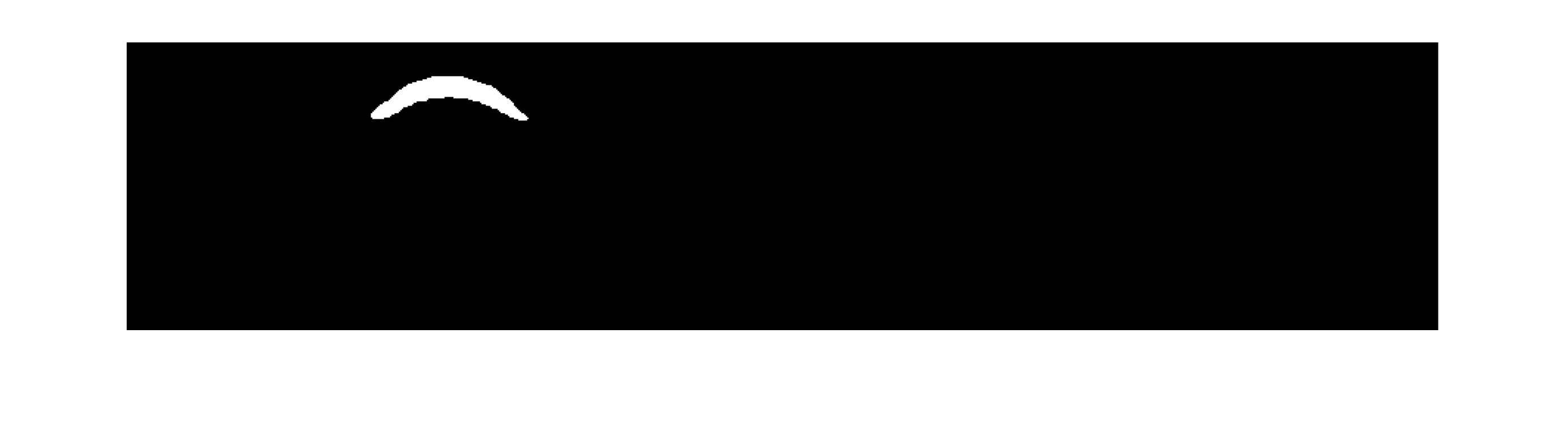}}
	\subfigure[]{ \centering
		\label{ellfit21}
		\includegraphics[width=0.20\textwidth]{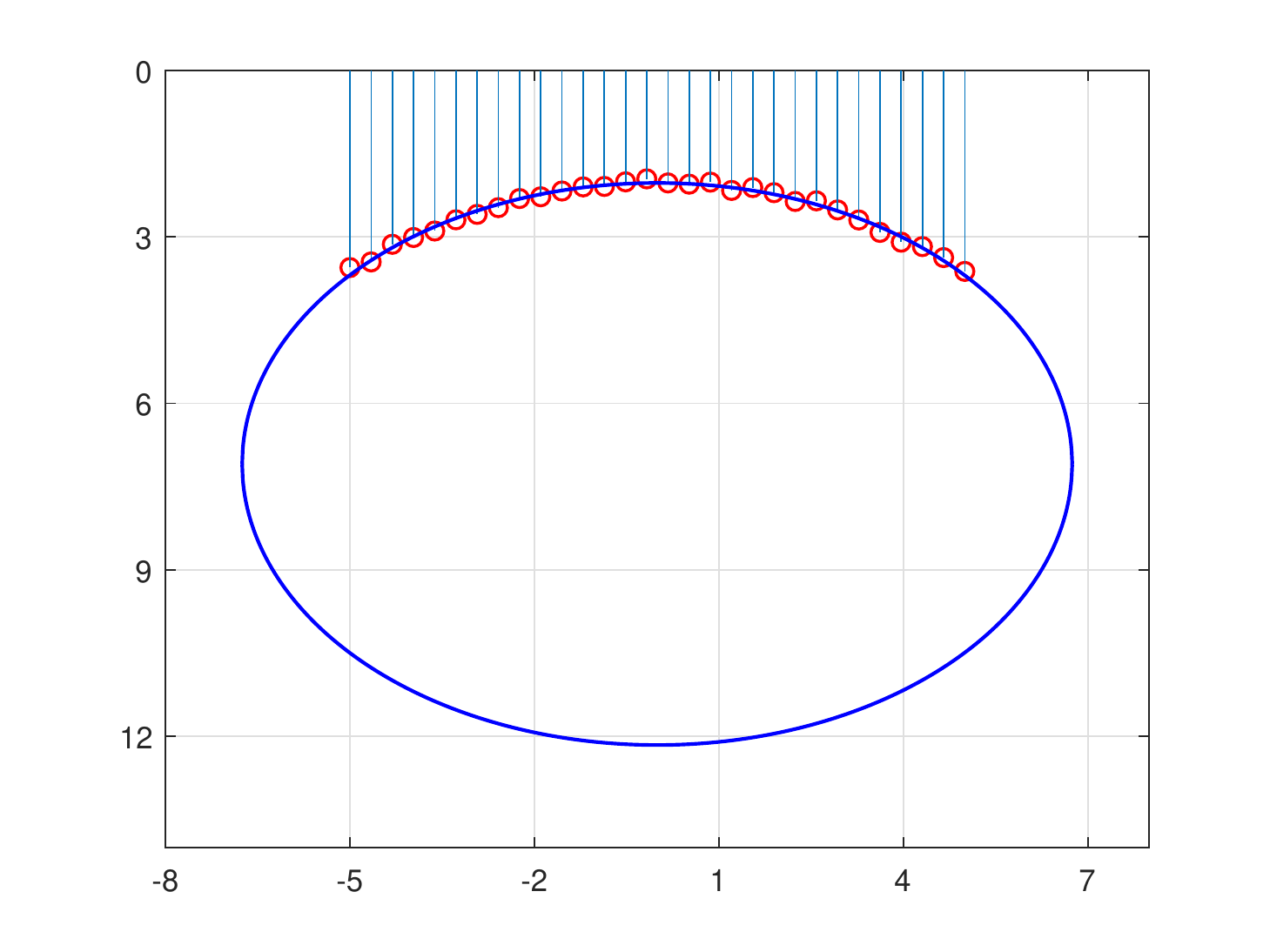}}
	\subfigure[]{ \centering
		\label{ellfit22}
		\includegraphics[width=0.20\textwidth]{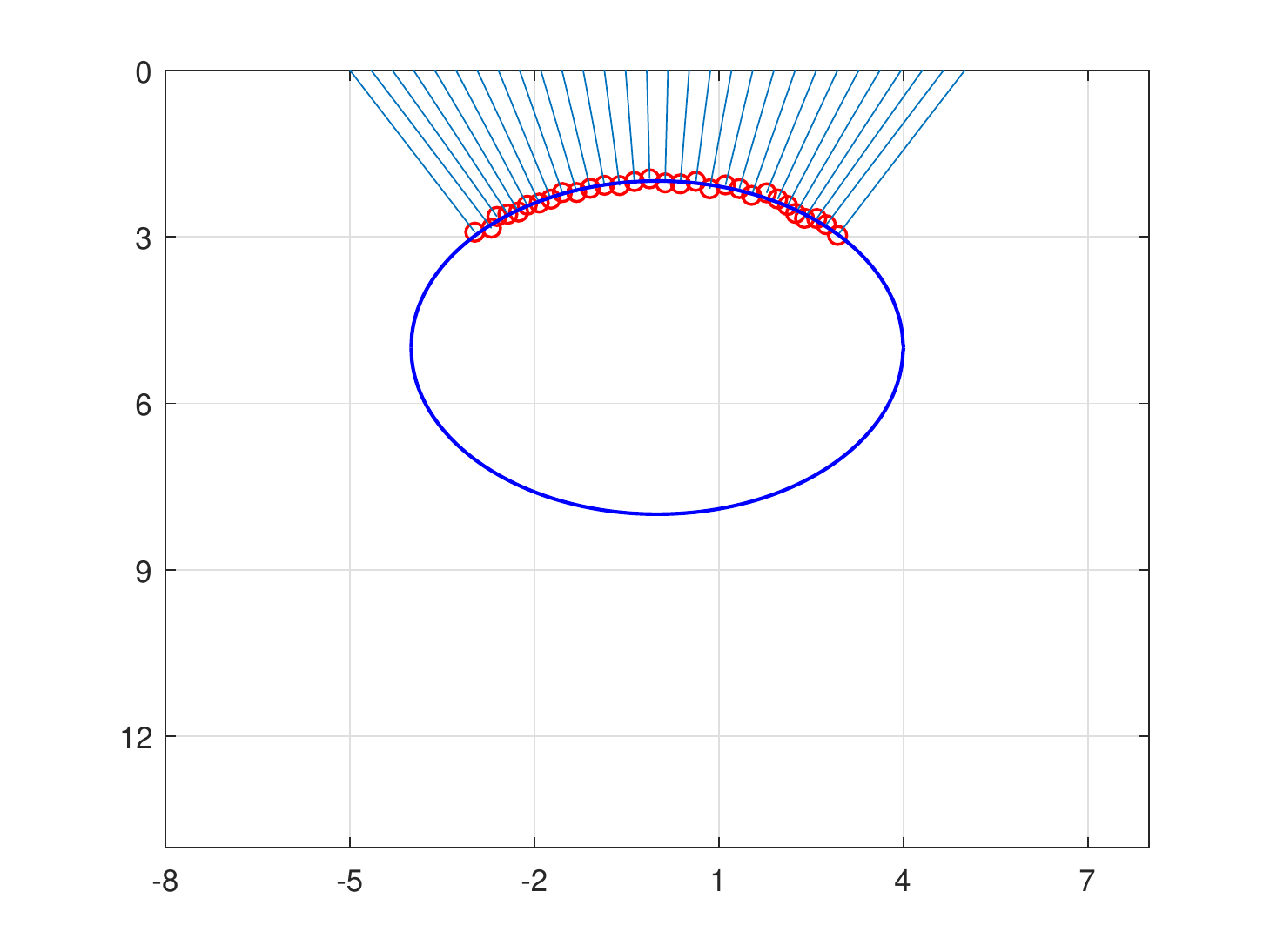}}
	\caption{The processing flow of the proposed model. (a) is the B-scan image. (b) is the obtained result after preprocessing and OSCA. (c) is the extracted point set and the fitted ellipse at the beginning of the iteration. (d) is the result of the proposed model, where the extracted points are inverted to the elliptical cross section of the pipe.}	\label{Gprimage}
\end{figure}

\begin{figure}[htbp]
	\centering
	\subfigure[]{ \centering
		\label{gprdirection11}
		\includegraphics[width=0.21\textwidth]{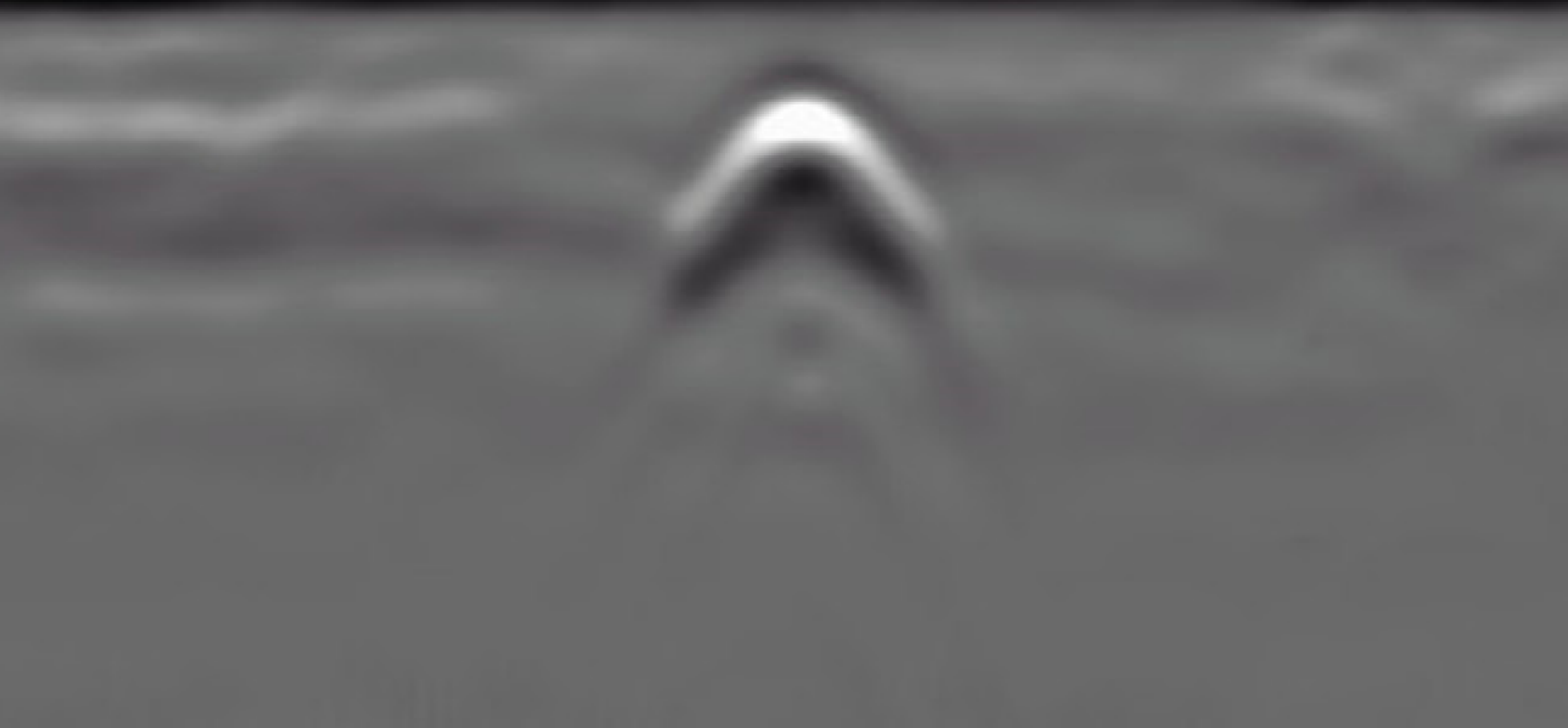}}
	\subfigure[]{ \centering
		\label{gprdirection12}
		\includegraphics[width=0.21\textwidth]{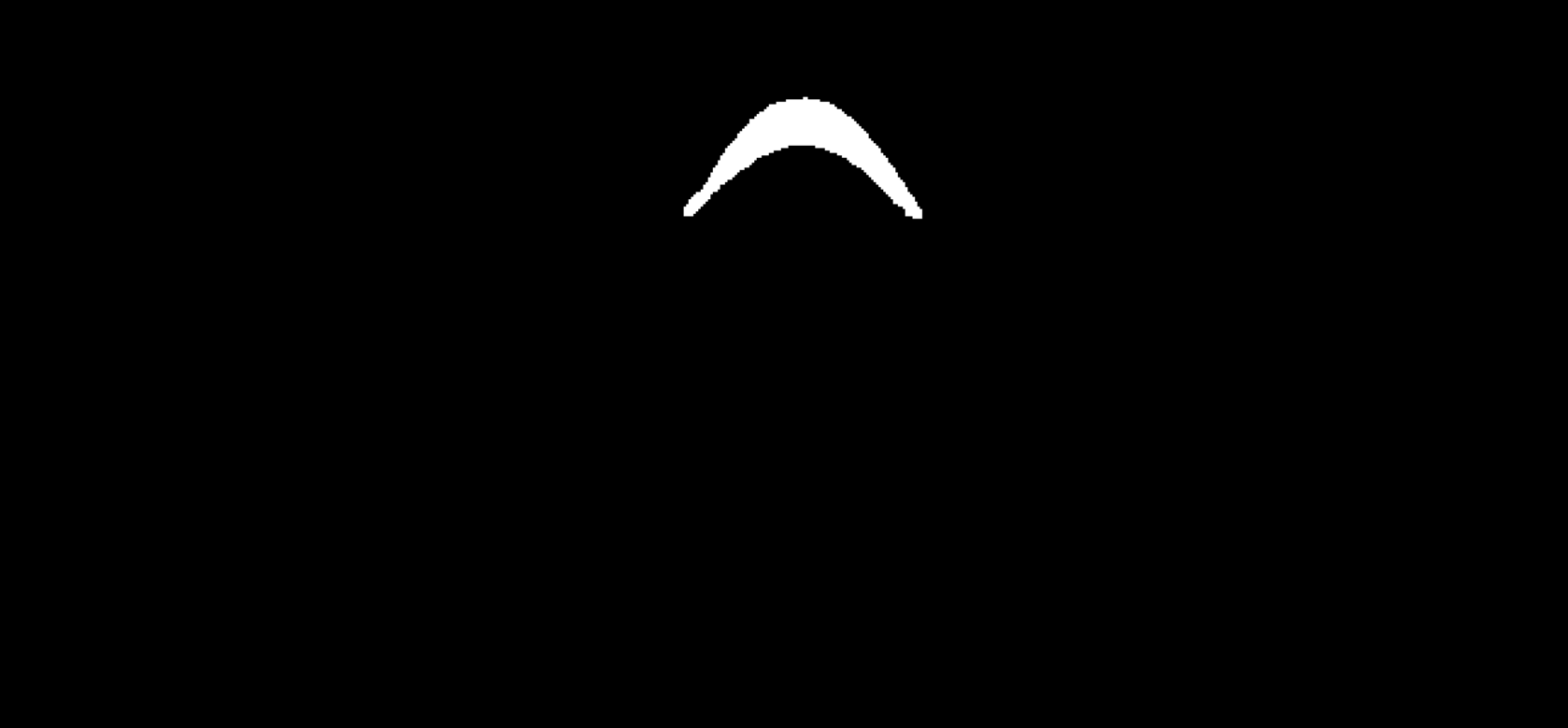}}
	\subfigure[]{ \centering
		\label{ellfit11}
		\includegraphics[height=1.55in]{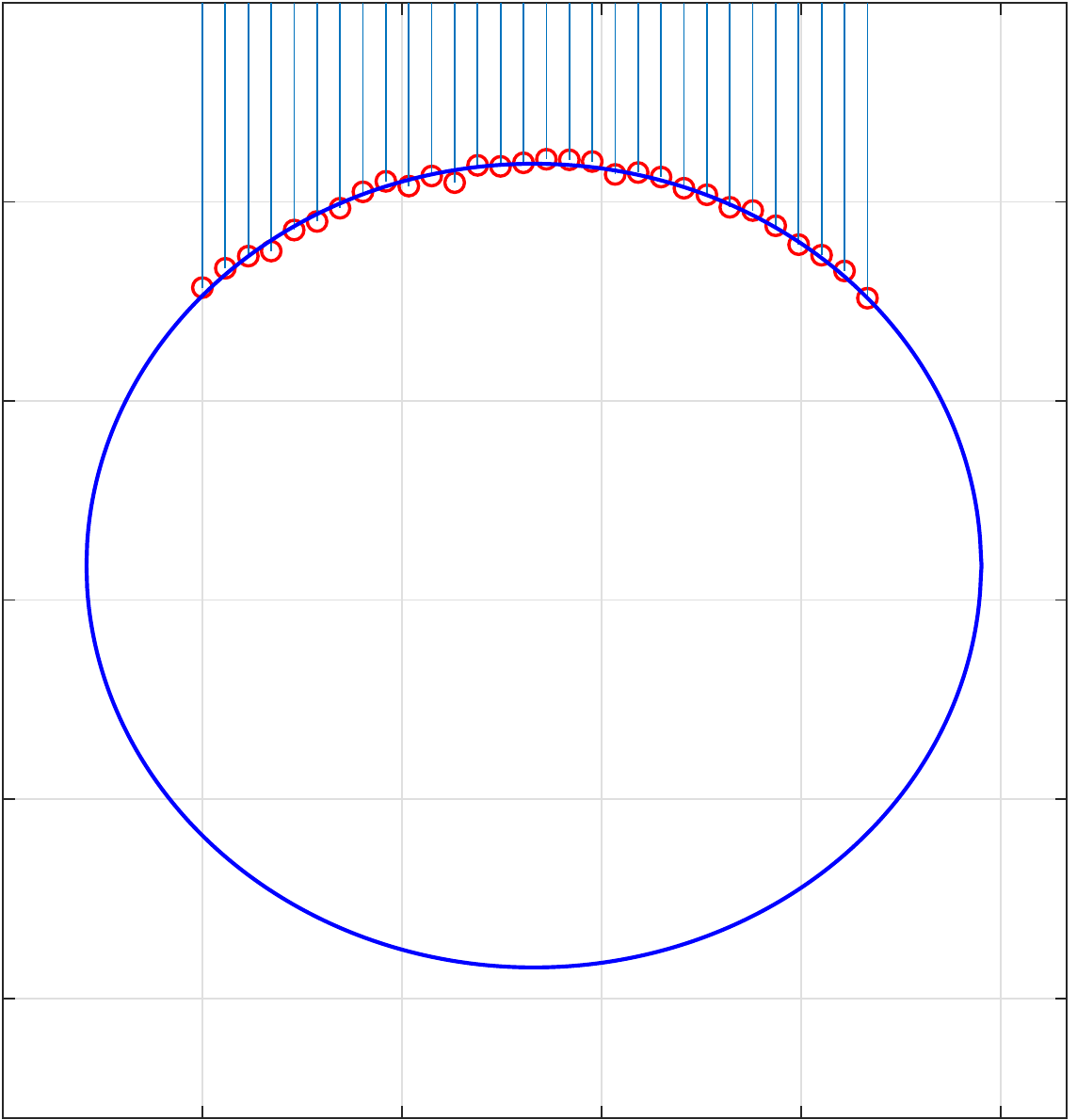}}
	\subfigure[]{ \centering
		\label{ellfit12}
		\includegraphics[height=1.55in]{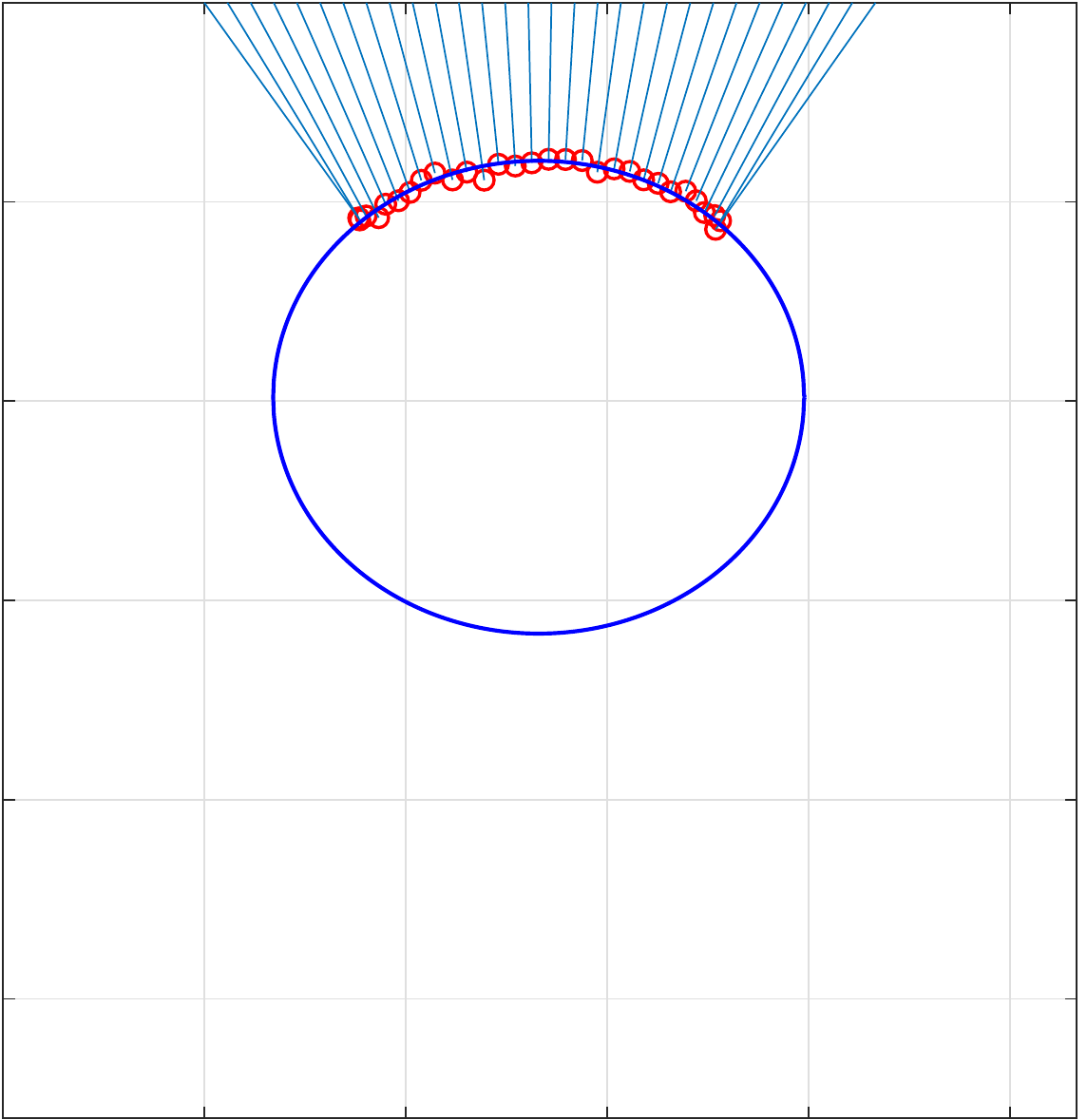}}
	\caption{These four pictures indicate another example of the proposed model, and the meaning of each picture is the same as the above one.}	\label{Gprimage2}
\end{figure}

Due to the limitation of the paper's length, the obtained GPR B-scan images could not be fully demonstrated here, and two images that illustrate the process of the proposed model are shown in Fig. \ref{Gprimage} and Fig. \ref{Gprimage2}. When extracting point sets from the obtained GPR B-scan images, the horizontal interval between each two points is $2$cm, and for each pipe, $30$ points are extracted. The maximum number of iterations $K$ is set to $10$, and the threshold of the sum of algebraic distance $D_t$ is set to $90$cm (the average fitting error of each extracted point is less than $3$cm). Specific analysis of the results are presented in the next subsection.

\subsection{Analysis of the experimental results}

\begin{figure}[htbp]
	\centering
	\subfigure[]{ \centering
		\label{modifymap1}
		\includegraphics[height=1.1in]{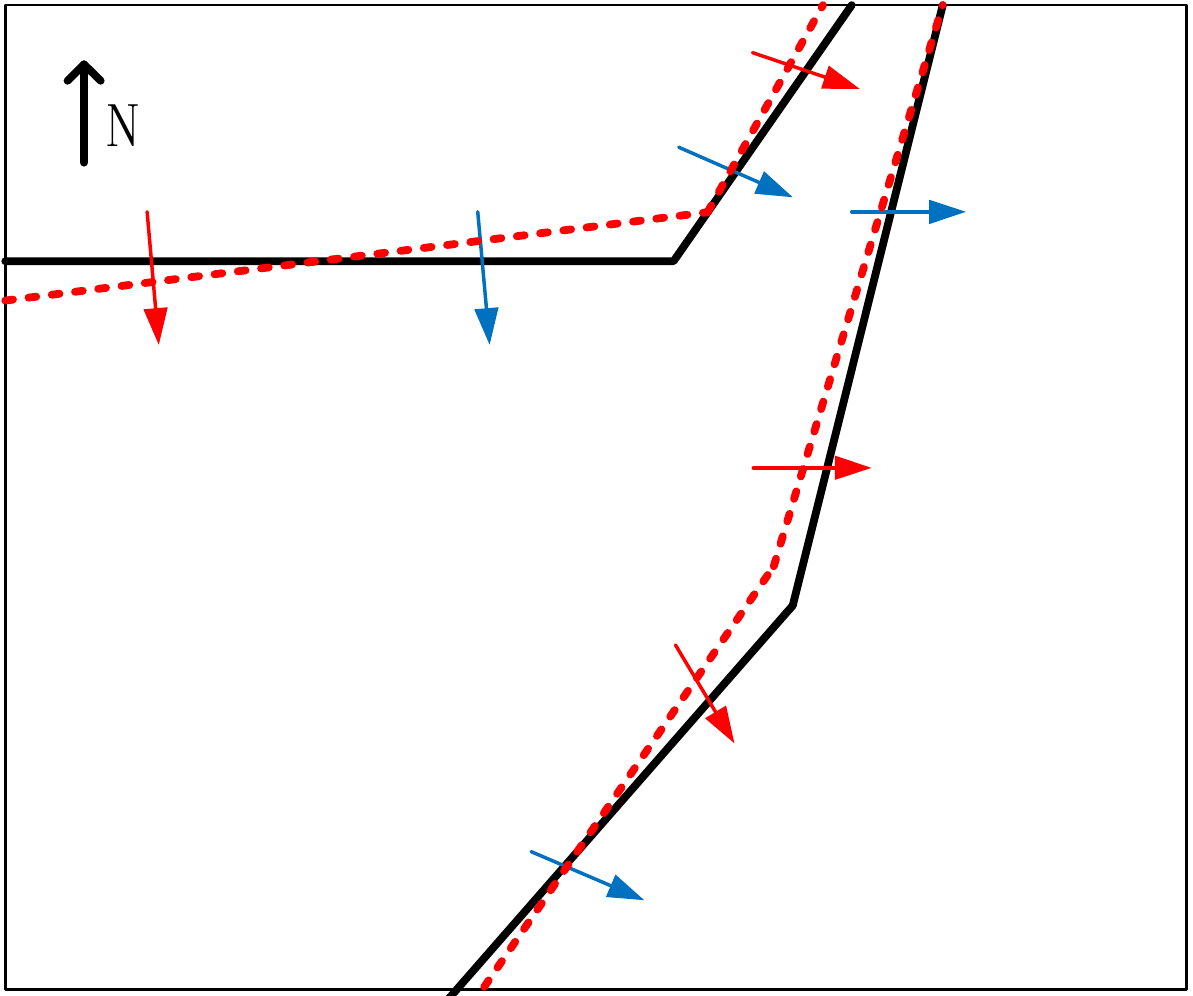}}
	\subfigure[]{ \centering
		\label{modifymap2}
		\includegraphics[height=1.1in]{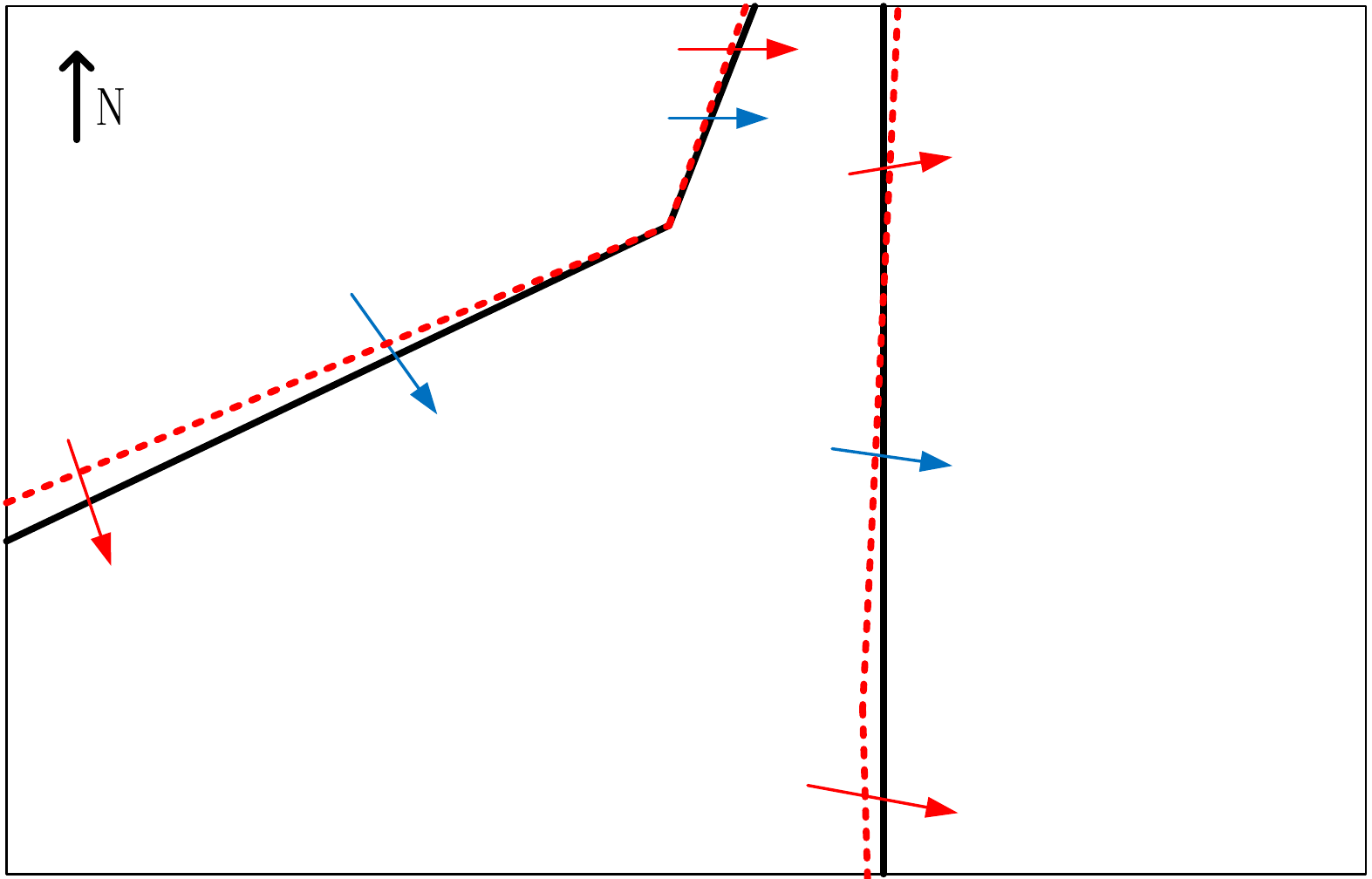}}
	\caption{More detections are conducted as the red lines with arrows in (a) and (b). The modified pipeline maps are shown as the red dotted lines.}	\label{modifymap}
\end{figure}

By applying the proposed model, the direction of buried pipes are obtained, by which the existing pipeline maps are modified as Fig .\ref{modifymap}. To validate the results, more detections are conducted as Fig. \ref{modifymap}, and evacuations are conducted to determine the actual direction and radius of each pipe. The average errors of Ellipse Inversion Model are presented in Table \ref{error}.
\begin{table}[htbp]
	\footnotesize
	\centering
	\caption{The errors of the proposed model}
	\begin{tabular}{ccccc}
		\toprule
		\multirow{2}[4]{*}{Area} & \multicolumn{2}{c}{The  average error } & \multicolumn{2}{c}{The max error} \\
		\cmidrule{2-5}          & Directions($\alpha$) &  Radius($b$) & Directions($\alpha$) &  Radius($b$) \\
		\midrule
		1     & 4.03\% & 5.23\% & 7.14\% & 7.22\% \\
		\midrule
		2     & 5.10\% & 5.90\% & 8.40\% & 7.41\% \\
		\bottomrule
	\end{tabular}%
	\label{error}%
\end{table}

As the result shows, the error of direction obtained by the proposed model could be controlled at about $5\%$ in the experimental environments of this paper. This is of great practical value in real-world applications, since only one detection is needed to confirm the general direction of the pipeline, which provides a basis for detection at the next position.
In the experimental settings, the maximum number of iterations is set to $10$. In the conducted detections, all extracted point sets converge and launch iterations within $10$ times.
Moreover, when the permittivity of the soil is known, the radius of the pipeline could be obtained through a detection that is not strictly required to be perpendicular to the pipeline. The Restricted Algebraic-Distance-based Fitting algorithm (RADF)\cite{zhou2018automatic} and Orthogonal-Distance-based Fitting with Constraints (ODFC) method\cite{dou2016real} are also applied to fit the extracted point sets to estimate the radius of the pipe, and the results are shown in Table \ref{errorfit}. It could be seen that when the GPR's detecting direction is not perpendicular to the pipeline, fitting the generated features by hyperbolic equations would leads to larger errors than the proposed model.

\begin{table}[htbp]
	\small
	\centering
	\caption{The average  error of radius by RADF, ODF and EIIA }
	\begin{tabular}{cccc}
		\toprule
		\multirow{2}[4]{*}{Area} & \multicolumn{3}{c}{The average  error of radius } \\
		\cmidrule{2-4}          & ODF   & RADF  & EIIA  \\
		\midrule
		1     & 15.20\% & 13.25\% & 5.23\% \\
		\midrule
		2     & 17.31\% & 15.59\% & 5.90\% \\
		\bottomrule
	\end{tabular}%
	\label{errorfit}%
\end{table}%

\section{Conclusion}

In this paper, a novel method to estimate the direction and radius of the buried pipeline from GPR B-scan image is proposed. The model consists of two parts: GPR B-scan image processing and Ellipse iterative inversion algorithm.
The GPR B-scan image is firstly processed with downward-opening point set extracted. Then the obtained point set is iteratively inverted to the cross section of the buried pipe, that is, the elliptical cross section caused by the angle between the GPR detection directions and the pipe direction. By minimizing the sum of the algebraic distances from these points to the inverted ellipse, the most likely pipe direction and radius are determined. 
Experiments on real-world datasets are conducted, and the existing pipeline map is modified, which validated the effectiveness of the proposed model. In future work, we will study how to map the pipelines of an area where there is no existing pipeline map.

\bibliography{ref}
\bibliographystyle{IEEEtran}

\end{spacing}
\end{document}